\documentclass{article}
\usepackage{mathptmx}
\usepackage[T1]{fontenc}
\usepackage[utf8]{inputenc}
\usepackage{amsmath}
\usepackage{amssymb}
\usepackage{graphicx}
\usepackage{microtype}

\makeatletter

\usepackage{amsthm}

 \usepackage[nonatbib,preprint]{nips_2018}

\usepackage{url}
\usepackage{booktabs}
\usepackage{amsfonts}
\usepackage{nicefrac}

\usepackage{caption}
\usepackage{color}
\usepackage{xcolor}

\title{Frequency Principle in Deep Learning with General Loss Functions and Its Potential Application}

\author{
  Zhi-Qin John Xu\thanks{This work is done while Xu is a visiting member at Courant Institute of Mathematical Sciences, New York University, New York, United States.} \\
  New York University Abu Dhabi\\
  Abu Dhabi 129188, United Arab Emirates \\
  \texttt{zhiqinxu@nyu.edu} \\
}

\@ifundefined{showcaptionsetup}{}{%
 \PassOptionsToPackage{caption=false}{subfig}}
\usepackage{subfig}
\makeatother

\begin{document}

\maketitle 
\begin{abstract}
Previous studies have shown that deep neural networks (DNNs) with
common settings often capture target functions from low to high frequency,
which is called Frequency Principle (F-Principle). It has also been
shown that F-Principle can provide an understanding to the often observed
good generalization ability of DNNs. However, previous studies focused
on the loss function of mean square error, while various loss functions
are used in practice. In this work, we show that the F-Principle holds
for a general loss function (e.g., mean square error, cross entropy,
etc.). In addition, DNN's F-Principle may be applied to develop numerical
schemes for solving various problems which would benefit from a fast
converging of low frequency. As an example of the potential usage
of F-Principle, we apply DNN in solving differential equations, in
which conventional methods (e.g., Jacobi method) is usually slow in
solving problems due to the convergence from high to low frequency. 
\end{abstract}

\section{Introduction}

Deep neural networks (DNNs) has achieved many state-of-the-art results
in various fields \cite{lecun2015deep}, such as object recognition,
language translation and game-play. A fully understanding of why DNNs
can achieve such good results remains elusive. The often used DNNs
equip much more parameters than the number of the training data. As
Von Neumann said “With four parameters I can fit an elephant, and
with five I can make him wiggle his trunk”. It is no surprise that
such DNNs can well fit the training data. However, counter-intuitive
to the traditional learning theory, such DNNs often do not overfit
(DNNs often generalize well to the test data which are not seen during
the training), which is often referred to as ``apparent paradox''
\cite{zhang2016understanding}.

A series of recent works \cite{xu_training_2018,xu2018understanding,rahaman2018spectral,arpit2017closer,poggio2018theory,perez2018deep,jakubovitz2018generalization,wu2017towards},
both experiments and theories, have gained us more understanding to
this paradox. In this work, we focus on the Fourier analysis of DNNs
\cite{xu_training_2018,xu2018understanding,rahaman2018spectral}.
Although the often used dataset, such as MNIST and CIFAR, are relative
simple compared with practical dataset, the input dimension (the pixel
number of each input image) is still very high for a quantitative
analysis. A good starting point to understand this apparent paradox
is to find an example that is simple enough for analysis but also
preserves this interesting paradox. The simple example turns out to
be the fitting of a function with one-dimension (1-d) input and 1-d
output \cite{xu_training_2018}. A prompt example to understand the
apparent paradox is that a very high-order polynomial fitting for
randomly sampled data points often overfit the training data, that
is, high oscillation occurs around the sample boundary (Runge's phenomenon);
however, a DNN with small weight initialization, no matter how large
size the DNN is, often learns the training data with a relative flat
function \cite{wu2017towards}. Starting from such 1-d functions,
experimentally \cite{xu_training_2018,rahaman2018spectral} and theoretically
\cite{xu2018understanding}, there exists a \emph{Frequency Principle}
(F-Principle) that DNNs often first quickly capture low-frequency
components while keeping high-frequency ones small, and then relatively
slowly captures high-frequency components. By F-Principle, the high-frequency
components of the DNN output is controlled by the training data. High
oscillation which exists in the Runge's phenomenon is then absent
in the DNN fitting. F-Principle also holds well in the often used
dataset \cite{xu_training_2018}, that is, MNIST and CIFAR-10. Theoretical
work indicates that the key ingredient underlying the F-Principle
in general DNN fitting problems is that the power spectrum of the
activation function decays in the Fourier space, where the power-decay
property is easy to be satisfied, such as sigmoid function and rectified
linear unit.

Previous studies focused on the DNN with mean square error \cite{xu_training_2018,xu2018understanding,rahaman2018spectral}.
It is yet to study whether F-Principle applies in the DNN with other
types of loss functions. This is important since loss function varies
in different problems, such as image classification and solving differential
equations \cite{weinan2018deep}. In this work, we perform a theoretical
analysis to show that for a general loss function, e.g., cross entropy,
the F-Principle qualitative holds in the DNN training, which is also
verified by experiments. The first experiment is a classification
problem with the loss function of cross entropy. The second experiment
is to apply DNN to solve Poisson equation by using Dirichlet's principle.

The DNN is a powerful tool to solve differential equations \cite{weinan2018deep,khoo2017solving,weinan2017deep},
especially for high-dimensional problems. It is well-known that different
frequencies converge with different speeds in solving differential
equations by numerical schemes. For example, for the Jacobi method,
low frequency converges much slower than high frequency. Multigrid
method is designed to speed up the convergence, which explicitly first
captures low-frequency parts \cite{briggs2000multigrid}. In addition,
manual frequency marching from low frequency to high frequency has
achieved great success in designing numerical schemes in various problems,
such as inverse scattering problems \cite{bao2015inverse} and Cryo-EM
reconstruction problems \cite{barnett2017rapid}. By showing F-Principle
in solving Poisson's equations, we emphasize that the DNN structure,
which implicitly endows low frequency with high priority, could be
a powerful tool to the problems that benefit from a fast converging
of low frequency. For example, we propose an ideal that combines DNN
and conventional methods (e.g., Jacobi method or Gauss-Seidel method),
in which DNN is in charge of capturing low-frequency parts and conventional
methods are in charge of capturing high-frequency parts. This idea
is exemplified by solving a 1-d Poisson's equation.

\section{F-Principle with general loss function}

Consider a general DNN, and denote its output as $\Upsilon_{\theta}(x)$,
where $\theta$ stands for the DNN parameters and $x$ stands for
the input. Represent $\Upsilon_{\theta}(x)$ with orthonormal basis
$\{p_{k}(x)\}$:

\begin{equation}
\Upsilon_{\theta}(x)=\sum_{k}c_{\theta,k}p_{k}(x),
\end{equation}
where $c_{\theta,k}$ is the coefficient of mode $k$ depending on
$\theta$. Denote the loss at sample $x$ as 
\begin{equation}
l_{x}=l(\Upsilon_{\theta}(x)).
\end{equation}
The total loss $L$ is 
\begin{equation}
L=\sum_{x}l_{x}=\sum_{x}l(\Upsilon_{\theta}(x)).
\end{equation}
Consider the gradient of the total loss with respect to parameter
$\theta$: 
\begin{align}
\frac{\partial L}{\partial\theta} & =\sum_{x}\frac{\partial l(\Upsilon_{\theta}(x))}{\partial\theta}\\
 & =\sum_{x}\frac{\partial l(\Upsilon_{\theta}(x))}{\partial\Upsilon}\frac{\partial\Upsilon_{\theta}(x)}{\partial\theta}\\
 & =\sum_{x}\frac{\partial l(\Upsilon_{\theta}(x))}{\partial\Upsilon}\sum_{k}p_{k}(x)\frac{\partial c_{\theta,k}}{\partial\theta}\\
 & =\sum_{k}\frac{\partial c_{\theta,k}}{\partial\theta}\sum_{x}\frac{\partial l(\Upsilon_{\theta}(x))}{\partial\Upsilon}p_{k}(x).
\end{align}
Let 
\begin{equation}
d_{k}\triangleq\sum_{x}\frac{\partial l(\Upsilon_{\theta}(x))}{\partial\Upsilon}p_{k}(x).
\end{equation}
$d_{k}$ is the coefficient of $\partial l(\Upsilon_{\theta}(x))/\partial\Upsilon$
at the component of $p_{k}$. Consider that $\{p_{k}(x)\}$ is Fourier
basis. According to Riemann-Lebesgue lemma, if a function is an integrable
function on an interval, then the Fourier coefficients of this function
tend to 0 as the order $k$ tends to infinity. Therefore, when the
activation function and the target function both are integrable functions
on the considered interval, $c_{\theta,k}$ and $d_{k}$ tend to 0
as the order $k$ tends to infinity. Denote 
\begin{equation}
\mathcal{L}_{k}\triangleq\sum_{k}\frac{\partial c_{\theta,k}}{\partial\theta}d_{k}.
\end{equation}
We have 
\begin{equation}
\frac{\partial L}{\partial\theta}=\sum_{k}\mathcal{L}_{k}.
\end{equation}
Therefore, we can decompose $\partial L/\partial\theta$ into a summation
of $\mathcal{L}_{k}$, which tends to 0 as the order $k$ tends to
infinity. This analysis implies that for any loss function, the change
of any parameter $\theta$ at each training step is affected more
by lower frequencies, which would rationalize the F-Principle in general
loss functions, as examined in the following experiments.

\section{Experiment: cross entropy loss}

The loss function of cross entropy is widely used in classification
problems. We use experiments to show that F-Principle holds in the
DNN training with this loss function.

\subsection{Toy data}

Consider a target function $y(x)=(y_{1}(x),y_{2}(x))$, where 
\begin{equation}
y_{1}(x)=\begin{cases}
1 & x\geq0\\
0 & x<0
\end{cases},
\end{equation}
\begin{equation}
y_{2}(x)=\begin{cases}
0 & x>0\\
1 & x\leq0
\end{cases}.
\end{equation}
This fitting problem is a toy classification problem. In the DNN in
this problem, the output layer has two neurons with softmax as activation
function. The output is denoted as $\Upsilon(x)=(\Upsilon_{1}(x),\Upsilon_{2}(x))$.
The loss function is 
\begin{equation}
L=\sum_{j=1}^{2}L_{j}=-\sum_{j=1}^{2}\sum_{x}y_{j}(x)\log\Upsilon_{j}(x)+(1-y_{j}(x))\log(1-\Upsilon_{j}(x)).
\end{equation}

For illustration, we focus on $y_{1}(x)$, which is shown in Fig.\ref{fig:Poisson-2}a.
Next, we examine the convergence of different frequencies. In a finite
interval, the frequency components of a target function are quantified
by Fourier coefficients computed from Discrete Fourier Transform (DFT).
Note that because the frequency in DFT is discrete, we can refer to
a frequency component by its index instead of its physical frequency.
The Fourier coefficient of $y_{1}(x)$ for the $\gamma$-th frequency
component is denoted by $F[y_{1}](\gamma)$ (a complex number in general).
$|F[y_{1}](\gamma)|$ is the corresponding amplitude, where $|\cdot|$
denotes the absolute value. Note that we call $\gamma$ the \emph{frequency
index}. $|F[y_{1}](\gamma)|$ is shown in Fig. \ref{fig:Poisson-2}b.
To examine the convergence behavior of different frequency components
during the training of a DNN, we compute the relative difference of
the DNN output $\Upsilon_{1}(x)$ and $y_{1}(x)$ in frequency domain
at each recording step, i.e., 
\begin{equation}
\Delta_{F}(\gamma)=\frac{|F[\Upsilon_{1}](\gamma)-F[y_{1}](\gamma)|}{|F[\Upsilon_{1}](\gamma)|}.\label{eq:dfreq-1}
\end{equation}

During the training, $\Upsilon_{1}(x)$ captures $y_{1}(x)$ from
low to high frequency in a clear order, as shown in Fig.\ref{fig:Poisson-2}c. 
\begin{center}
\begin{figure*}
\begin{centering}
\subfloat[]{\begin{centering}
\includegraphics[scale=0.27]{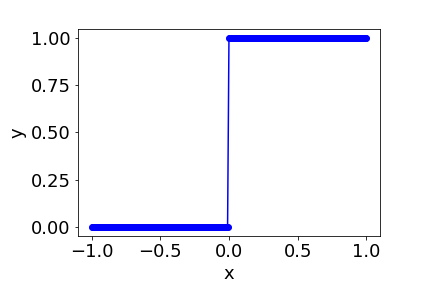} 
\par\end{centering}
}\subfloat[]{\begin{centering}
\includegraphics[scale=0.27]{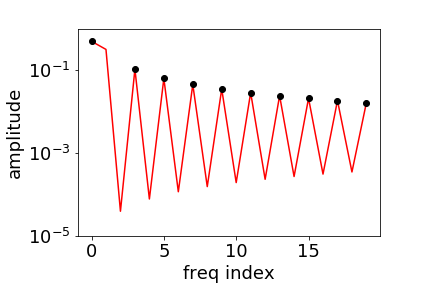} 
\par\end{centering}
}\subfloat[DNN ]{\begin{centering}
\includegraphics[scale=0.27]{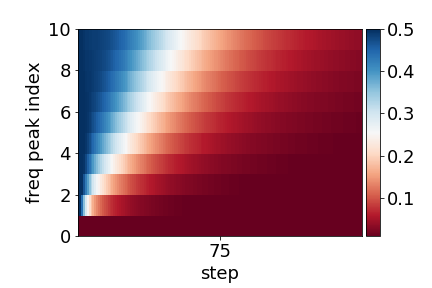} 
\par\end{centering}
}
\par\end{centering}
\caption{F-Principle with cross entropy loss. The first output dimension of
the target function is shown in (a) and its Fourier coefficient amplitude
as a function of frequency index is shown in (b). Frequency peaks
are marked by black dots. (c) Relative difference at different recording
steps for different selected frequency indexes. The training data
are evenly sampled in $[-1,1]$ with sample size $201$. We use a
DNN with width 400-400-200-100 with full batch training, the output
layer has two neurons with softmax as activation function, and learning
rate is $2\times10^{-4}$. The parameters of the DNN are initialized
following a Gaussian distribution with mean $0$ and standard deviation
$0.1$.\label{fig:Poisson-2} }
\end{figure*}
\par\end{center}

\subsection{MNIST data}

To verify that the F-Principle holds in the image classification problems
(MNIST) with the loss function of cross entropy, we perform Fourier
analysis in the first principle component of the input space. The
procedure is as follows.

The training set is a list of images with labels: $\{\vec{x}_{k},\vec{y}_{k}\}_{k=0}^{n-1}$.
Each image is represented by a vector $\vec{x}_{k}\in\mathbb{R}^{N_{in}}$,
where $N_{in}=784$ is the pixel number of an image. $\vec{y}_{k}$
is an one-hot vector indicating the label. The dimensions for the
input layer and the output layer\emph{ }are $N_{in}$ and $10$, respectively.
First, we compute the first principle direction. Transform each image
by 
\begin{equation}
\vec{x}_{j}^{\prime}=\vec{x}_{j}-\frac{1}{n}\sum_{k=0}^{n-1}\vec{x}_{k},\quad j=0,1,\cdots,n-1.
\end{equation}
Denote all images by $X=[\vec{x}_{0}^{\prime},\vec{x}_{1}^{\prime},\cdots,\vec{x}_{n-1}^{\prime}]\in\mathbb{R}^{N_{in}\times n}$.
The covariance matrix is $C_{x}=XX^{T}$. The eigenvector of the maximal
eigenvalue of $C_{x}$ can be obtained, denoted by $\vec{p}_{1}\in\mathbb{R}^{N_{in}}$,
i.e., the first principle direction. The projection of each image
in the $\vec{p}_{1}$ direction is $x_{k}^{\prime}\triangleq\vec{p}_{1}^{T}\vec{x}_{k}$.
We rescale $x_{k}^{\prime}$ to $x_{k}$ such that $x_{k}\in[0,1]$
by 
\begin{equation}
x_{k}=\frac{(x_{k}^{\prime}-\min_{j}x_{j}^{\prime})}{\max_{l}(x_{l}^{\prime}-\min_{j}x_{j}^{\prime})}.
\end{equation}
Then, the sample set is $S=\{(x_{0},\vec{y}_{0}),(x_{1},\vec{y}_{1}),\cdots,(x_{n-1},\vec{y}_{n-1})$.
For illustration, we only consider the first component of $\vec{y}$,
i.e., $\vec{y}^{(1)}$. Note that now $\{x_{k}\}_{k=0}^{n-1}$ is
a non-uniform sampling. Using non-uniform FFT (NUFFT), we can obtain
\begin{equation}
F[\vec{y}^{(1)}][\gamma_{k}]=\sum_{j=0}^{n-1}\vec{y}_{k}^{(1)}\exp\left(-2\pi ix_{j}k\right).
\end{equation}
Then, the sampling on the Fourier domain is 
\begin{equation}
S_{\gamma}=\{(\gamma_{0},F[\vec{y}^{(1)}](\gamma_{0})),(\omega_{1},F[\vec{y}^{(1)}](\gamma_{1})),\cdots,(\gamma_{n-1},F[\vec{y}^{(1)}](\gamma_{n-1}))\}.
\end{equation}
After each training step, we feed each $\vec{x}_{k}$ into the DNN
and obtain the DNN output $\Upsilon(\vec{x}_{k})$: 
\begin{equation}
S_{T}=\{(x_{0},\Upsilon(\vec{x}_{0})),(x_{1},\Upsilon(\vec{x}_{1})),\cdots,(x_{n-1},\Upsilon(\vec{x}_{n-1}))\}.
\end{equation}
Using non-uniform FFT (NUFFT), similarly, we can obtain the sampling
of the DNN's first dimension output on the Fourier domain 
\begin{equation}
S_{\Upsilon,\gamma}=\{(\gamma_{0},F[\Upsilon^{(1)}](\gamma_{0})),(\omega_{1},F[\Upsilon^{(1)}](\gamma_{1})),\cdots,(\gamma_{n-1},F[\Upsilon^{(1)}](\gamma_{n-1}))\},
\end{equation}
which is shown in Fig.\ref{fig:Poisson-2-1}a. We then examine the
relative error of certain selected important frequency components
(marked by black squares). As shown in the first column in Fig.\ref{fig:Poisson-2-1}b,
we can observe that the DNN tends to capture low-frequency components
first. 
\begin{center}
\begin{figure*}
\begin{centering}
\subfloat[]{\begin{centering}
\includegraphics[scale=0.27]{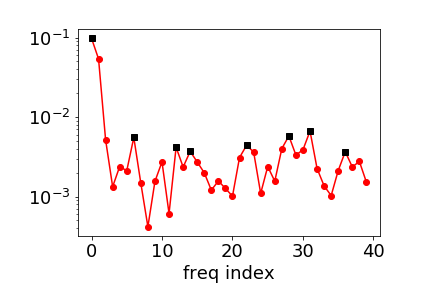} 
\par\end{centering}
}\subfloat[]{\begin{centering}
\includegraphics[scale=0.27]{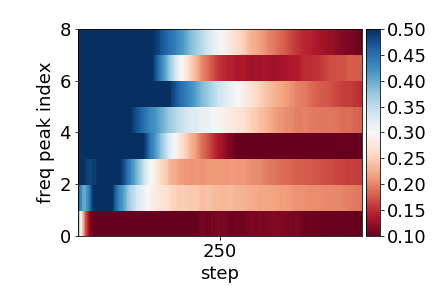} 
\par\end{centering}
}
\par\end{centering}
\caption{F-Principle with cross entropy loss on MNIST dataset. The Fourier
coefficient amplitude of the first output dimension of the target
function is shown in (a). Frequency peaks are marked by black dots.
(b) Relative difference at different recording steps for different
selected frequency indexes. The training data are $10000$ test samples
of MNIST dataset. We use a DNN with width 400-200 with batch size
as 128, the output layer has 10 neurons with softmax as activation
function, and learning rate is $10^{-5}$. The parameters of the DNN
are initialized following a Gaussian distribution with mean $0$ and
standard deviation $0.2$. \label{fig:Poisson-2-1} }
\end{figure*}
\par\end{center}

\section{Experiment: Poisson's equations}

Consider one-dimension (1-d) Poisson's equation \cite{weinan2018deep,evans2010partial}:
\begin{equation}
-\triangle u(x)=g(x),\quad x\in\Omega=(-1,1)\label{eq:Poisson1}
\end{equation}
\[
u(x)=0,\quad x=-1,1.
\]
The Poisson's equation can be solved by numerical schemes (e.g., Jacobi
method) or DNN. As well known, high frequency converges faster in
the Jacobi method. In the following, we would show that high frequency
converges slower when the DNN is applied to solve the above Poisson's
equation.

\subsection{Central differencing scheme and Jacobi method}

$[-1,1]$ is uniformly discretized into $n+1$ points with step $\Delta x=2/n$,
i.e., $x_{0},x_{1},\cdots,x_{n}$. The Poisson's equation in Eq. (\ref{eq:Poisson1})
can be solved by central differencing scheme: 
\begin{equation}
-\triangle u_{i}=-\frac{u_{i+1}-2u_{i}+u_{i-1}}{(\triangle x)^{2}}=g(x_{i}),\quad i=1,2,\cdots,n.
\end{equation}
Write the above in the matrix form: 
\begin{equation}
Au=g,\label{eq:auf}
\end{equation}
where 
\begin{equation}
A=\left(\begin{array}{cccccc}
2 & -1 & 0 & 0 & \cdots & 0\\
-1 & 2 & -1 & 0 & \cdots & 0\\
0 & -1 & 2 & -1 & \cdots & 0\\
\vdots & \vdots & \cdots &  &  & \vdots\\
0 & 0 & \cdots & 0 & -1 & 2
\end{array}\right)_{(n-1)\times(n-1)},
\end{equation}

\begin{equation}
u=\left(\begin{array}{c}
u_{1}\\
u_{2}\\
\vdots\\
u_{n-2}\\
u_{n-1}
\end{array}\right),\quad g=(\triangle x)^{2}\left(\begin{array}{c}
g_{1}\\
g_{2}\\
\vdots\\
g_{n-2}\\
g_{n-1}
\end{array}\right),\quad x_{i}=2\frac{i}{n}.
\end{equation}
If $n$ is not a large number, Eq. (\ref{eq:auf}) can be solved by
performing the inverse of $A$. When $n$ is a very large number,
this problem can be solved by iterative schemes. For example, we illustrate
the Jacobi method. Let $A=D-L-U$, where $D$ is diagonal, and $L$
and $U$ are the strictly lower and upper parts of $-A$, respectively.
Then, we can obtain 
\begin{equation}
u=D^{-1}(L+U)u+D^{-1}g.
\end{equation}
The Jacobi iteration is 
\begin{equation}
u^{l+1}=D^{-1}(L+U)u^{l}+D^{-1}g.
\end{equation}
We perform error analysis of the above iteration process. Denote $u^{*}$
as the true value obtained by directly performing inverse of $A$
in Eq. (\ref{eq:auf}). The error at step $l+1$ is $e^{l+1}=u^{l+1}-u^{*}$.
Then, $e^{l+1}=R_{J}e^{l}$, where $R_{J}=D^{-1}(L+U)$. The converging
speed of $e^{l}$ is determined by the eigenvalues of $R_{J}$, that
is, 
\begin{equation}
\lambda_{k}=\lambda_{k}(R_{J})=\cos\frac{k\pi}{n},\quad k=1,2,\cdots,n-1,
\end{equation}
and the corresponding eigenvector $v_{k}$ is 
\begin{equation}
v_{k,j}=\sin\frac{jk\pi}{n},j=1,2,\cdots,n-1.
\end{equation}
Write 
\begin{equation}
e^{l}=\sum_{k=1}^{n-1}\alpha_{k}^{l}v_{k},
\end{equation}
where $\alpha_{k}^{l}$ can be understood as the magnitude of $e^{l}$
in the direction of $v_{k}$. Then, 
\begin{equation}
e^{l+1}=\sum_{k=1}^{n-1}\alpha_{k}^{l}R_{J}v_{k}=\sum_{k=1}^{n-1}\alpha_{k}^{l}\lambda_{k}v_{k}.
\end{equation}
\[
\alpha_{k}^{l+1}=\lambda_{k}\alpha_{k}^{l}.
\]
Therefore, the converging speed of $e^{l}$ in the direction of $v_{k}$
is controlled by $\lambda_{k}$. Since 
\begin{equation}
\cos\frac{k\pi}{n}=-\cos\frac{(n-k)\pi}{n},
\end{equation}
the frequencies $k$ and $(n-k)$ are closely related and converge
with the same speed. Consider the frequency $k<n/2$, $\lambda_{k}$
is larger for lower frequency. Therefor, lower frequency converges
slower in the Jacobi method.

\subsection{DNN approach}

Similar as the loss function in Ref \cite{weinan2018deep}, we consider
the following loss function (energy method) 
\begin{equation}
I(u)=\int_{\Omega}\left(\frac{1}{2}|\nabla_{x}u(x)|^{2}-g(x)u(x)\right)dx+\beta\int_{\partial\Omega}u(x)^{2}ds.\label{eq:Energy}
\end{equation}
It is equivalent to solve Poisson's equation by finding the function
that minimizes $I(u)$ (Dirichlet’s principle) \cite{evans2010partial}.
The last term in $I(u)$ is a penalty in order to satisfy the boundary
condition. The DNN structure is 1-d input (i.e., $x$) and 1-d output
(denoted as $\Upsilon(x)$) for solving Eq. (\ref{eq:Poisson1}).
$\beta$ is a constant.

The procedure is similar. We discretized $[-1,1]$ into $n+1$ even-space
points. In each training step, we compute $\Upsilon(x_{i})$ for $i=0,1,2,\cdots n$.
The gradient of $I(\Upsilon)$ with respect to parameter $\Theta$
is 
\begin{equation}
\frac{dI(\Upsilon)}{d\Theta}=\sum_{i=0}^{n}\frac{dI(\Upsilon(x_{i}))}{d\Theta}.
\end{equation}
At each training step, we compare $\Upsilon(x)$ and $u^{*}(x)$ in
the Fourier domain. Note that $u^{*}(x)$ is the one obtained by directly
performing inverse of $A$ in Eq. (\ref{eq:auf}).

\subsection{Experiment}

Consider 
\begin{equation}
g(x)=\sin(x)+4\sin(4x)-8\sin(8x)+16\sin(24x).
\end{equation}
As shown in Fig. \ref{fig:Poisson}a, after training, the DNN output
can well fit the solution $u^{*}$ obtained by directly performing
inverse of $A$ in Eq. (\ref{eq:auf}). As shown in Fig. \ref{fig:Poisson}b,
there are three peaks of $u^{*}$ in the Fourier domain.

To examine the convergence behavior of different frequency components
during the DNN training, we compute the relative difference of the
DNN output $\Upsilon(x)$ and $u^{*}(x)$ in frequency domain at each
recording step, i.e., 
\begin{equation}
\Delta_{F}(\gamma)=\frac{|F[u^{*}](\gamma)-F[\Upsilon](\gamma)|}{|F[u^{*}](\gamma)|}.\label{eq:dfreq}
\end{equation}
As shown in Fig. \ref{fig:Poisson}c, F-Principle holds well in solving
Poisson's equation \cite{xu_training_2018,xu2018understanding}. For
comparison, we also show that low frequency converges much slower
than high frequency in Jacobi method, as shown in Fig. \ref{fig:Poisson}d. 
\begin{center}
\begin{figure}
\begin{centering}
\subfloat[]{\begin{centering}
\includegraphics[scale=0.27]{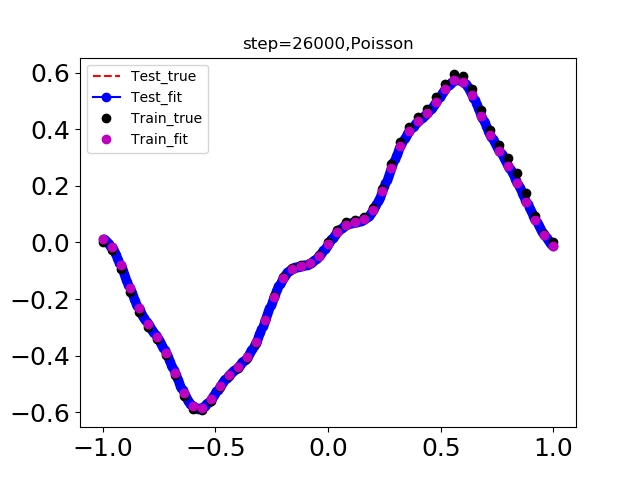} 
\par\end{centering}
}\subfloat[]{\begin{centering}
\includegraphics[scale=0.27]{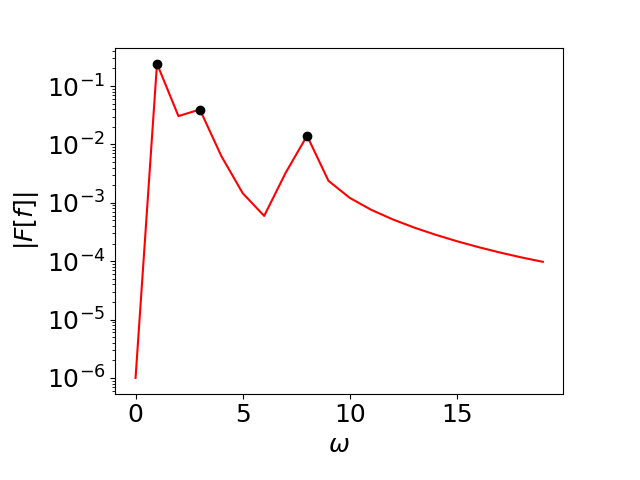} 
\par\end{centering}
}
\par\end{centering}
\begin{centering}
\subfloat[DNN ]{\begin{centering}
\includegraphics[scale=0.27]{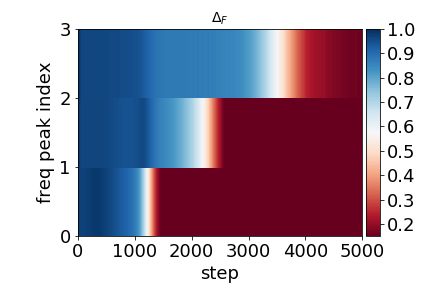} 
\par\end{centering}
}\subfloat[ Jacobi]{\begin{centering}
\includegraphics[scale=0.27]{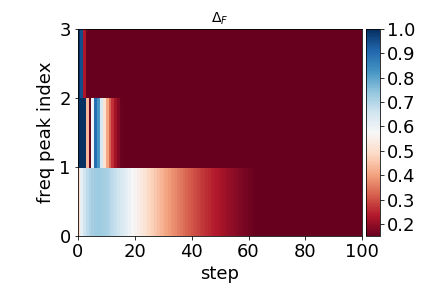} 
\par\end{centering}
}
\par\end{centering}
\caption{Frequency domain analysis of the Poisson's equation in Eq. (\ref{eq:Poisson1})
with $g(x)=\sin(x)+4\sin(4x)-8\sin(8x)+16\sin(24x)$. (a) $u(x)$.
The true value is computed by central differencing scheme with directly
compute the inverse of coefficient matrix. (b) $|F[u^{*}]|$ (red
solid line) as a function of frequency index. Frequency peaks are
marked by black dots. (c, d) Relative difference at different recording
steps for different selected frequency indexes. The training data
and the test data are evenly sampled in $[-1,1]$ with sample size
$51$ and $401$, respectively. We use a DNN with width 4000-800 with
full batch training. The learning rate is $5\times10^{-6}$ at beginning
and halved every $10000$ training epochs. $\beta$ is $10$. Each
step consists of four epochs. (d) is the result of Jacobi iteration.
The parameters of the DNN are initialized following a Gaussian distribution
with mean $0$ and standard deviation $0.05$. \label{fig:Poisson} }
\end{figure}
\par\end{center}

\section{Combination of DNN and convention methods}

In light of the above numerical simulations, it is natural to consider
if we can combine DNN and Jacobi method to solve the Poisson's equation.
For simplicity, we call the combination method D-Jacobi method.

In the first part of D-Jacobi method, we solve the Poisson's equation
by DNN with $M$ steps. In the second part, we use the DNN output
at step $M$ as the initial value for the Jacobi method.

We solve the problem in Fig. \ref{fig:Poisson} by a laptop (Dell,
Precision 5510). As shown in Fig.\ref{fig:Poisson-1}a, the DNN loss
fluctuates after some running time. We use Jacobi method to solve
the problem after some time points, which are indicated by vertical
dashed lines. As shown in Fig.\ref{fig:Poisson-1}b (Fig.\ref{fig:Poisson-1}c),
green stars indicates the $|\Upsilon-u^{*}|_{\infty}$ of the DNN
output at different steps. Dashed lines indicates the evolution of
the Jacobi (Gauss-Seidel) method. As we can see, if the selected timing
is too early, it would still take long time to converge to a small
error, because the low frequencies are not converged, yet. If the
selected timing is too late, much time would be waste because the
DNN is hard to capture high frequencies and fluctuates a lot. The
selected timing of the green or the red one is a better choice. In
practice, a better way to select the timing is when the loss gets
flat and fluctuated for a short while. 
\begin{center}
\begin{figure*}
\begin{centering}
\subfloat[Loss]{\begin{centering}
\includegraphics[scale=0.3]{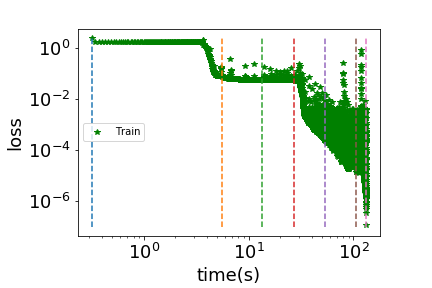} 
\par\end{centering}
}\subfloat[Jacobi: $|u-u^{*}|_{\infty}$]{\begin{centering}
\includegraphics[scale=0.3]{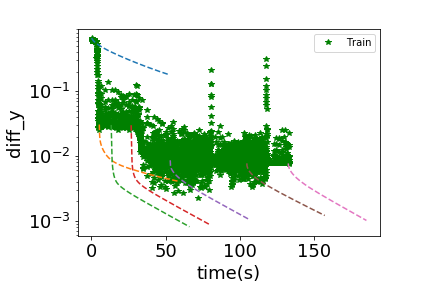} 
\par\end{centering}
}\subfloat[GS: $|u-u^{*}|_{\infty}$]{\begin{centering}
\includegraphics[scale=0.3]{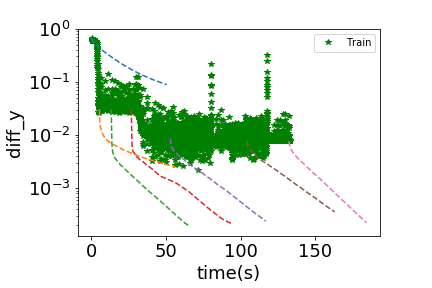} 
\par\end{centering}
}
\par\end{centering}
\caption{Combination methods for solving the Poisson's equation in Eq. (\ref{eq:Poisson1})
with $g(x)=\sin(x)+4\sin(4x)-8\sin(8x)+16\sin(24x)$. The abscissa
is the computer running time. (a) Loss is the form in Eq. (\ref{eq:Energy}).
We use Jacobi method to solve the problem after several time points,
which are indicated by vertical dashed lines. (b) Green stars indicates
the $|\Upsilon-u^{*}|_{\infty}$ at different steps. Dashed lines
indicates the evolution of the Jacobi method. (c) Gauss-Seidel method.
The training data are evenly sampled in $[-1,1]$ with sample size
$1001$. $u^{*}$ is computed by central differencing scheme with
directly computing the inverse of coefficient matrix. We use a DNN
with width 4000-500-400 with full batch training, and learning rate
is $5\times10^{-4}$. $\beta$ is $10$. The parameters of the DNN
are initialized following a Gaussian distribution with mean $0$ and
standard deviation $0.02$.\label{fig:Poisson-1} }
\end{figure*}
\par\end{center}

\section{Discussion}

In this work, we have shown that F-Principle holds well in the DNN
training with a general loss function, extending the study of F-Principle
in the loss function of mean square error in previous works \cite{xu_training_2018,xu2018understanding,rahaman2018spectral}.
Along with the previous study that F-Principle holds in both DNN and
convolutional neural networks with the activation function of either
tanh or Relu \cite{xu_training_2018,xu2018understanding}, these works
implicate that the F-Principle may provide understandings to the generalization
ability of general DNNs.

We also show that the generality of F-Principle in the DNN training
could potentially be useful in designing algorithms for solving practical
problems. To be specific, we apply DNN to solve 1-d Poisson's equation.
Compared with conventional numerical schemes, DNN could potentially
work better in rather high dimensions \cite{weinan2018deep}. In addition,
it does not requires discretization for the DNN method, which would
be much easier to be implemented. In future, it would be interested
to use DNN's F-Principle to develop numerical schemes for solving
various problems which would benefit from a fast converging of low
frequency.

\section*{Acknowledgments}

The author wants to thank Weinan E, Wei Cai for helpful discussion.
The author also wants to thank Tao Luo, Zheng Ma, Yanyang Xiao and
Yaoyu Zhang for the discussion of the F-Principle. This work was funded
by the NYU Abu Dhabi Institute G1301.

\subsection*{ \bibliographystyle{unsrt}
\bibliography{C:/GDrive/Research/DeepLearning-xzx/BIB/DLRef}
 }
\end{document}